%% file: InteractiveLearningObjectProperties.tex
\documentclass[letterpaper, 10 pt, conference]{ieeeconf}  % Comment this line out if you need a4paper
                             
\usepackage{graphicx}
\usepackage{multirow}
\usepackage{amsmath}
\usepackage{amssymb}
\usepackage{csquotes}
\usepackage{url}
\usepackage{hyperref}
\usepackage{subcaption}
\usepackage{color, colortbl}
\usepackage[ruled,linesnumbered]{algorithm2e}

\usepackage[nolist]{acronym} % acronyms by the \ac{label} command
\let\labelindent\relax
\usepackage[inline]{enumitem} % inline enumerate
\usepackage{cite}
\graphicspath{ {Figures/} }
\usepackage{siunitx}
\newif\ifshowComments
\showCommentstrue
\ifshowComments
    \usepackage[draft,markup=bfit, deletedmarkup=sout, authormarkup=brackets]{changes}
\else
    \usepackage[final,markup=bfit, deletedmarkup=sout, authormarkup=brackets]{changes}
\fi

\usepackage[firstpageonly=true]{draftwatermark}

\definechangesauthor[name={Lukas},color=green]{LR}
\definechangesauthor[name={Matej}, color=orange]{MH}
\definechangesauthor[name={Andrej}, color=yellow]{AK}
\definechangesauthor[name={Shubhan}, color=black]{SPP}
\definechangesauthor[name={Jan}, color=black]{JB}
\definechangesauthor[name={Jiri}, color=purple]{JH}

\ifshowComments

\else

\fi

\input{acronyms.tex}
\input{macro.tex}

\useVspacetrue % true to use vspaces

\title{Interactive Learning of Physical Object Properties Through Robot Manipulation and Database of Object Measurements}

\author{Andrej Kruzliak$^{1}$, Jiri Hartvich$^{1}$, Shubhan P. Patni$^{1}$, Lukas Rustler$^{1}$, Jan Kristof Behrens$^{2}$, Fares J. Abu-Dakka$^{3}$, \\Krystian Mikolajczyk$^{4}$, Ville Kyrki$^{5}$, and Matej Hoffmann$^{1}$
\thanks{$^{1}$Department of Cybernetics, Faculty of Electrical Engineering, Czech Technical University in Prague; {\tt\small matej.hoffmann@fel.cvut.cz} }
\thanks{$^{2}$Czech Institute of Informatics, Robotics, and Cybernetics, Czech Technical University in Prague}
\thanks{$^{3}$Mechanical Engineering Department, Faculty of Engineering, New York University Abu Dhabi, UAE. Part of the work was done in Mondragon Unibertsitatea}
\thanks{$^{4}$Department of Electrical and Electronic Engineering, Imperial College London, London, UK}
\thanks{$^{5}$Intelligent Robotics Group, Department of Electrical Engineering and Automation, School of Electrical Engineering, Aalto University}
    \thanks{This work was co-funded by the European Union under the project ROBOPROX (reg. no. CZ.02.01.01/00/22\_008/0004590). S.P.P. and L.R. were additionally supported by the Grant Agency of the CTU in Prague (no. SGS24/096/OHK3/2T/13).
    This work originated in the IPALM project (H2020-FET-ERA-NET Cofund-CHIST-ERA III / TAČR EPSILON, no. TH05020001). We would like to thank Tran Nguyen Le for the first code prototype capturing the relationships between some objects' properties. }
}

\IEEEoverridecommandlockouts
\begin{document}

\SetWatermarkAngle{0}
\SetWatermarkColor{black}
\SetWatermarkLightness{0.5}
\SetWatermarkFontSize{9pt}
% \SetWatermarkScale
% \SetWatermarkHorCenter
\SetWatermarkVerCenter{30pt}
\SetWatermarkText{\parbox{30cm}{%
\centering This is the authors' final version of the manuscript published as:\\
\centering A. Kruzliak et al., ``Interactive Learning of Physical Object Properties Through Robot Manipulation and Database of Object Measurements,'' \\
\centering in Intelligent Robots and Systems (IROS), IEEE/RSJ International Conference on, pp. 7596-7603, 2024, (C) IEEE, \\
\centering \url{https://doi.org/10.1109/IROS58592.2024.10802249} \\
\centering An accompanying video is available at \url{https://youtu.be/h_ZIYUmzv-8?si=obTHd1wen35Bdjdx}.
}}

\maketitle

%%%%%%%%%%%%%%%%%%%%%%%%%%%%%%%%%%%%%%%%%%%%%%%%%%%%%%%%%%%%%%%%%%%%%%%%%%%%%%%%
% RA-L 1800 characters
% IROS 3000 characters 
\begin{abstract}
This work presents a framework for automatically extracting physical object properties, such as material composition, mass, volume, and stiffness, through robot manipulation and a database of object measurements. The framework involves exploratory action selection to maximize learning about objects on a table. A Bayesian network models conditional dependencies between object properties, incorporating prior probability distributions and uncertainty associated with measurement actions. The algorithm selects optimal exploratory actions based on expected information gain and updates object properties through Bayesian inference. Experimental evaluation demonstrates effective action selection compared to a baseline and correct termination of the experiments if there is nothing more to be learned. The algorithm proved to behave intelligently when presented with trick objects with material properties in conflict with their appearance. The robot pipeline integrates with a logging module and an online database of objects, containing over 24,000 measurements of 63 objects with different grippers. All code and data are publicly available, facilitating automatic digitization of objects and their physical properties through exploratory manipulations.
\end{abstract}

\begin{figure}[ht]
	\centering
	\def\svgwidth{\linewidth}
	{\fontsize{9}{9}%\selectfont\sf
		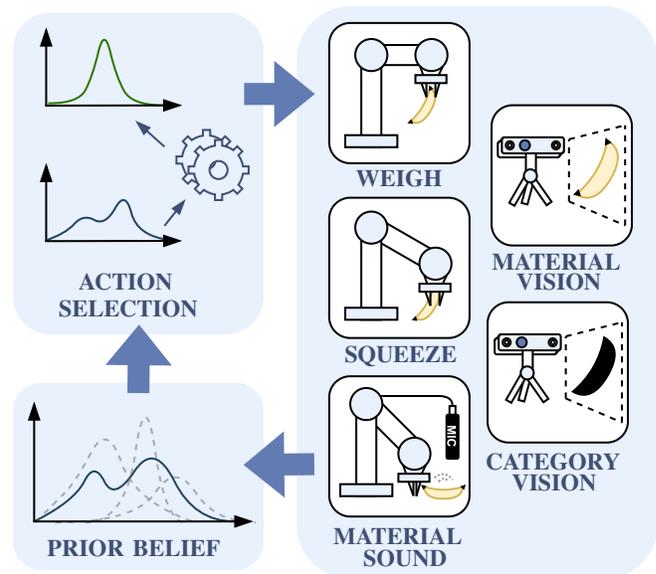}
	\caption{Exploratory action selection to estimate object properties -- schematic overview. Using the current belief, next action is selected to maximize the expected \acf{ig}, taking the uncertainty of each action into account. After each action, new belief is computed using inference in the Bayesian Network, and a new action is selected.}
	\label{fig:acsel_loop}
\customVspace{-1.5em}
\end{figure}

\section{Introduction}
%Extracting properties of objects from visual input only can be ambiguous, especially regarding physical properties like surface roughness, stiffness, or mass. Here, manipulation or haptic exploration is indispensable. 
%Furthermore, unlike visual sensing which is often passive from images taken by a static camera, haptic exploration is intrinsically active: the particular way of manipulating the object, like sequence of finger movements of a robot hand, determines the quality of information that can be acquired. 

While rapid progress is being made in automatically extracting information about objects from large image datasets and text corpora, estimating physical properties remains a challenge. This work paves the way to automatic extraction of physical object properties such as material composition, mass, volume or stiffness, which typically require physically manipulating the object. We present a theoretical framework and its experimental evaluation that deals with exploratory action selection to maximize what can be learned about objects on the table in front of a robot manipulator (see Fig.~\ref{fig:acsel_loop} for an overview). In our demonstrator, there is a partially adversarial mix of 17 household objects of different shapes and from different materials (incl. a real and a plastic banana, wine glass from glass and plastic, a soft and a hard sponge with identical appearance), 6 object properties (2 discrete -- category and material; 4 continuous -- elasticity, density, volume, and mass), and 5 actions the robot system can perform (2 based on vision -- estimate the category or material from the image; 3 manipulation actions -- squeeze, poke, and lift). We expressed the conditional dependencies between the properties using a \ac{bn} (see Fig.~\ref{fig:probability_model}) and added freely available prior probability distributions as well as the uncertainty associated with different measurement actions. After specifying which object property should be estimated, our algorithm chooses the best exploratory action based on expected \acf{ig}, updates all object properties using inference in the \ac{bn}, and picks the next optimal action to be executed. We demonstrate that the overall average performance of the system outperforms a baseline and analyze interesting runs showing how the algorithm automatically and efficiently discovers object properties. The robot pipeline is integrated with a logging module and an online database of objects, currently containing more than 24000 measurements of 63 different objects using four different grippers. We make all code and data publicly available, which provides a starting point for automatic digitization of objects and their physical properties by exploratory manipulations.

Task and motion planning for robot manipulation, including sequences of actions, has been extensively studied using symbolic and recently also neuro-symbolic or neural approaches. In this work, the goal is not to perform a task (e.g., stack objects) but to learn about the properties of objects. Embodied reasoning for discovering object properties via manipulation was used in \cite{behrens2021embodied}; correct execution of a verbally specified task (e.g., stack objects by weight) required specific actions to learn about physical object properties. In this work, the order of actions is chosen not to produce a cumulative result in the environment, but so that the action chosen next is optimal with respect to what can be learned about the object in front of the robot.

%\section{Related Work}

%\textbf{Exploratory action selection}.
Active manipulation for perception can be non-prehensile (e.g. ``push to know'' \cite{dutta2023push,le2021probabilistic}) or prehensile, i.e. grasping the object or exploring it with robot fingers (e.g., \cite{xu2013tactile}). In most cases, manipulation serves object pose estimation (e.g., \cite{murali2022empirical}), shape reconstruction, or recognition; learning physical object properties has been less studied (see \cite{li2020review,petrovskaya2016active} for reviews). Haptic exploration of physical object properties using several actions and sensory modalities is extremely rare (with \cite{chu2013using} a notable exception). The exploratory movements are typically chosen based on some form of probabilistic model and information theoretic measures like entropy or \ac{ig} (see e.g., \cite{fishel2012bayesian, nikandrova2014towards}).  
% alternatively \cite{dutta2023push}

An accompanying video is available at \url{https://youtu.be/h_ZIYUmzv-8};
the code used in this work is hosted at
\url{https://github.com/ctu-vras/actsel}; The database of object measurements and its source code can be found at \url{https://cmp.felk.cvut.cz/ipalm/}.

\section{Exploratory Action Selection Driven by Information Gain} 
\label{sec:methods_inference}
\begin{figure}[tb]
\includegraphics[width=\columnwidth]{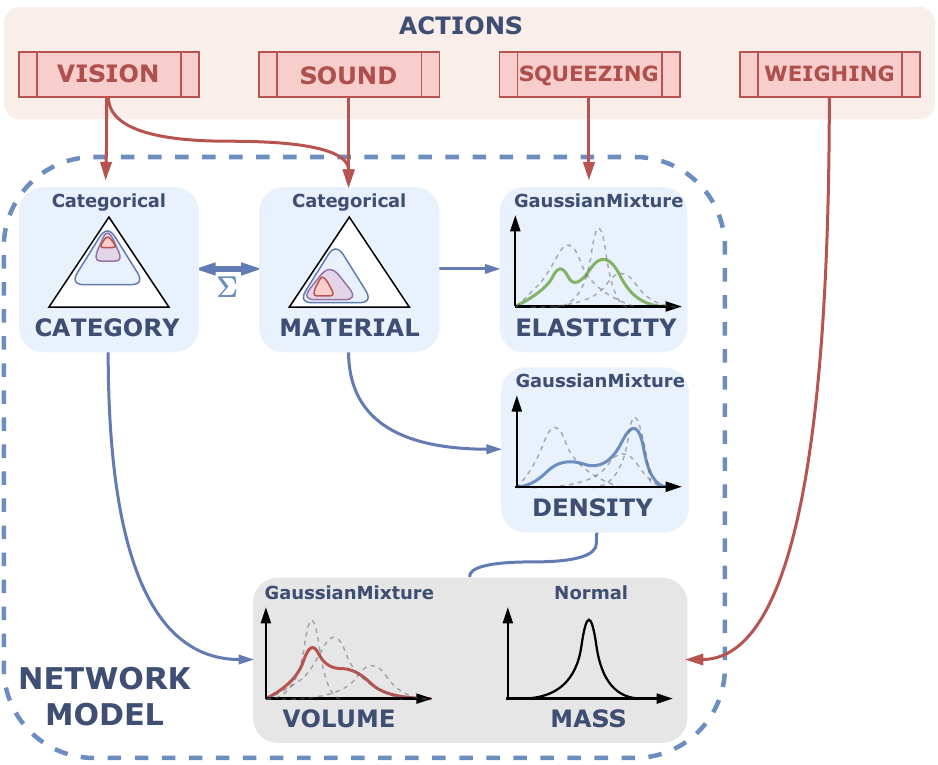}
\caption{Bayesian Network Structure.  }
\label{fig:probability_model}
\customVspace{-1em}
\end{figure}

Our method is based on a \acf{bn}. Beliefs about each of the variables can be updated through measurements. 
The action-to-node relation can be found in \figref{fig:probability_model}. 
The \ac{bn} consists of five nodes ordered by their enumerability types.
\begin{enumerate*}[label=(\roman*)]
    \item categorical (enumerable): \textit{Category} and \textit{Material};
    \item continuous (non-enumerable): \textit{Elasticity}, \textit{Density}, and \textit{Volume}.
\end{enumerate*}
%\textit{Category} assigns the object's category to a common household object classification, such as \textit{bottles} or \textit{bowls}, while \textit{Material} assigns the object's material property to a common material classification, such as \textit{ceramic} or \textit{glass}.
Discrete variables are represented with a \acf{pmf}; continuous variables with a \acf{pdf}.

We consider 10 object \textit{Categories}: \textit{bottle, bowl, box, can, dice, fruit, mug, plate, sodacan, wineglass}.

We consider 8 \textit{Material} classes: \textit{ceramic, glass, metal, soft plastic, hard plastic, wood, paper, foam.}

%Continuous properties are evaluated in the continuous space of the given property:
\textit{Elasticity} is a continuous variable expressed in Young's modulus [kPa], \textit{Density} in homogeneous density  [\textit{kg/$m^3$}], and \textit{Volume} in the volume of the object's material in [m$^3$]. 

%\begin{figure}[h]
%	\centering
%	\def\svgwidth{\linewidth}
%	{\fontsize{7}{7}%\selectfont\sf
%		\input{Figures/diagram_actions_probability_model_cropped.pdf_tex}}
%	\caption{Bayesian Network Structure.}
%	\label{fig:probability_model}
%	\customVspace{-1em}
%\end{figure}

%Each \textit{Material} is associated with a reference normal Gaussian distribution, with mean $\mu$ and variance $\sigma^2$, for \textit{Elasticity} and \textit{Density}, sourced from reference technical documents~\cite{mahendran_1996, dondi1999, HULAN2020, giancoli_2014}.

\subsection{Initialization}
\label{sec:init_bayes_values}

%The reference is used for each property except the volume node. That is because the volume is dependent on the category object only statistically, not naturally. In other words, the volume of material used to create a mug has a certain distribution for pragmatic reasons, but nothing stops one from making an arbitrarily big or small mug, therefore using larger or smaller volumes of ceramic. However, the density of the ceramic stays the same, ignoring the volume of the material used. It is, therefore, inherent. We assume inherence also for the elasticity because all of the elasticity measurements were conducted in regions of the objects with homogenous matter distribution, i.e., the volumetric properties of the objects did not play a role in the elasticity estimation.

\begin{figure}[tb]
\includegraphics[width=\columnwidth]{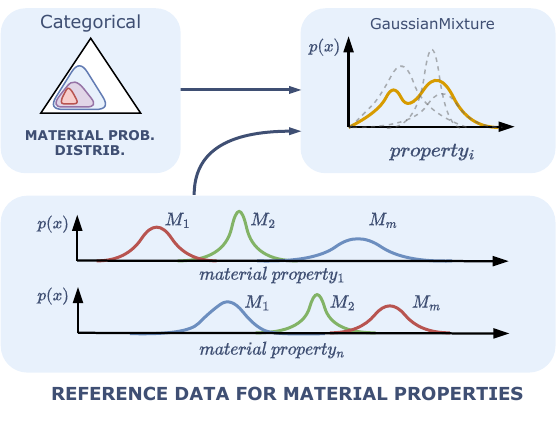}
\caption{Initialization of \ac{gmm} from the reference data -- weighting by corresponding probability mass function.}
\label{fig:gmm}
%\customVspace{-1.5em}
\end{figure}

To establish initial beliefs for each property, first, a uniform \ac{pmf} is set up for \textit{Category} and \textit{Material}. Those in turn initialize each continuous variable expressed by the relationships in the \ac{bn} (Fig.~\ref{fig:probability_model}), as shown in Fig.~\ref{fig:gmm}, giving rise to a \acf{gmm} for each continuous variable. 

The \textit{Category} \ac{pmf} is translated into the \textit{Volume} \ac{pdf} through the values in Table~\ref{tab:ref_volume}. The \textit{Material} \ac{pmf} is translated into the \textit{Elasticity} and \textit{Density} \acp{pdf} through the means and \acp{sd}  in Table~\ref{tab:ref_density_elasticity},  %Reference values (mean and \ac{sd}) for \textit{Density} and \textit{Elasticity} are necessary to initialize \acp{gmm} for these properties. 
extracted from technical literature~\cite{mahendran_1996, dondi1999, HULAN2020, giancoli_2014}. 

%, which This initializes \ac{gmm} for each property. The initialization of the \acp{gmm} is done by weighting each reference entry (see \tabref{tab:ref_density_elasticity}) for each property by the corresponding probability mass function---either associated with the \textit{Category} or the \textit{Material} node. This process is illustrated in \figref{fig:gmm}, resulting in the establishment of the initial belief about each property. 
%The mutual conditional dependence between \textit{Category} and \textit{Material} is represented by a covariance matrix, the specific values of which are discussed in \secref{sec:init_bayes_values}.  
%When this initialization is finished, inference can take place. 

\begin{table}
	\rmfamily\centering
	\resizebox{\linewidth}{!}{%
		\renewcommand\arraystretch{1.2} 
		\centering
		\begin{tabular}{|>{\arraybackslash}m{0.09\linewidth}|>{\centering\arraybackslash}m{0.12\linewidth}|>{\centering\arraybackslash}m{0.1\linewidth}|>{\centering\arraybackslash}m{0.1\linewidth}|>{\centering\arraybackslash}m{0.1\linewidth}|>{\centering\arraybackslash}m{0.13\linewidth}|}
			\noalign{\hrule height 1.5pt}
			\multirow{4}{*}{\begin{tabular}[c]{@{}c@{}}\textbf{Volume}\\ {[}cm$^3${]}\end{tabular}} & \textbf{Bottle} & \textbf{Bowl} & \textbf{Box} & \textbf{Can} & \textbf{Dice} 
			\\ \cline{2-6}
			& $30 \pm 50$ & $40 \pm 80$ & $10 \pm 26$ & $8 \pm 60$ & $37 \pm 50$ 
			\\ \cline{2-6} 
			& \textbf{Fruit} & \textbf{Mug} & \textbf{Plate} & \textbf{Sodacan} & \textbf{Wineglass} 
			\\ \cline{2-6} 
			& $155 \pm 28$ & $97 \pm 27$ & $0 \pm 15$ & $5 \pm 10$ & $85 \pm 20$ 
			\\ \hline
			\noalign{\hrule height 1.5pt}
		\end{tabular}%
	}
	\caption{Reference values for \textit{Volume}. The values are means and \acfp{sd} scraped from an online store.}
	\label{tab:ref_volume}
	\customVspace{-1em}
\end{table}

\begin{table}
	\rmfamily\centering
	\resizebox{\linewidth}{!}{%
		\renewcommand\arraystretch{1.2} 
		\centering
		\begin{tabular}{|>{\centering\arraybackslash}m{0.14\linewidth}|>{\centering\arraybackslash}m{0.15\linewidth}|>{\centering\arraybackslash}m{0.16\linewidth}|>{\centering\arraybackslash}m{0.19\linewidth}|>{\centering\arraybackslash}m{0.17\linewidth}|}
			\noalign{\hrule height 1.5pt}
			&\textbf{Ceramic}        & \textbf{Glass} & \textbf{Metal} & \textbf{Soft Plastic} 
			\\ \hline
			\textbf{Density} {[}kg$\cdot$ m$^{-3}${]} & $2300\pm 100$ & $2600\pm 200$ & $7900 \pm 600$   & $1300 \pm 300$ 
			\\ \hline
			\textbf{Elasticity} {[}kPa{]} & $150 \pm 10$  & $140 \pm 10$  & $130 \pm 10$   & $55 \pm 30$ 
			\\ \hline
			& \textbf{Hard Plastic} & \textbf{Wood} & \textbf{Paper} & \textbf{Foam} 
			\\ \hline
			\textbf{Density} {[}kg$\cdot$ m$^{-3}${]} & $1500 \pm 300$ & $646 \pm 183.6$ & $756.8 \pm 223.8$ & $91.9 \pm 127.7$
			\\ \hline
			\textbf{Elasticity} {[}kPa{]} & $75 \pm 30$ & $120 \pm 10$ & $35 \pm 28$ & $10 \pm 15$  
			\\ \hline
			\noalign{\hrule height 1.5pt}
		\end{tabular}%
	}
	\caption{Reference values for density and elasticity. The values are means and \acfp{sd} extracted from the literature.}
	\label{tab:ref_density_elasticity}
	%\customVspace{-1em}
\end{table}

%\begin{table}[]
%\resizebox{\columnwidth}{!}{%
%\begin{tabular}{|c|c|c|}
%\hline
%& \textbf{Density} [kg$\cdot m^{-3}$] & \textbf{Elasticity} [kPa] \\ \hline
%\textbf{Ceramic} & 2300 $\pm$ 100 & 150 $\pm$ 10 \\ \hline
%\textbf{Glass} & 2600 $\pm$ 200 & 140 $\pm$ 10 \\ \hline
%\textbf{Metal} & 7900 $\pm$ 600 & 130 $\pm$ 10 \\ \hline
%\textbf{Soft Plastic} & 1300 $\pm$ 300 & 55 $\pm$ 30 \\ \hline
%\textbf{Hard Plastic} & 1500 $\pm$ 300 & 75 $\pm$ 30 \\ \hline
%\textbf{Wood} & 646 $\pm$ 183.6 & 120 $\pm$ 10 \\ \hline
%\textbf{Paper} & 756.84 $\pm$ 223.76 & 35 $\pm$ 28 \\ \hline
%\textbf{Foam} & 91.95 $\pm$ 127.65 & 10 $\pm$ 15 \\ \hline
%\end{tabular}%
%}
%\caption{Reference values for density and elasticity. The values are means and \acfp{sd} extracted from the literature.}
%\label{tab:ref_density_elasticity}
%\customVspace{-1em}
%\end{table}

The conditional dependence between \textit{Category} and \textit{Volume} as well as the covariance matrix linking \textit{Category} and \textit{Material} were initialized using data extracted from the text descriptions of products available at an online store. %The values are available in \tabref{tab:ref_volume}. %Amazon

\subsection{Inference} 
\label{subsub:inference}
 Following the execution of an action, the network is updated based on the obtained results (measurement update). These data points are then incorporated as a product into the distribution of the given property. For categorical properties, the posterior distribution update is propagated to all other nodes in the \ac{bn}, adhering to the network's connections as an undirected graph. This updating process, known as message passing, involves transmitting message vectors between nodes. The format and order of these message vectors correspond to the relevant probability functions: for interactions between nodes related to \textit{Material}, the format aligns with the material \ac{pmf}, while for other interactions, it aligns with the category \ac{pmf}. 

When the message originates from a continuous property node, such as \textit{Elasticity}, the continuum must be first transformed into a finite vector to align with the required message format. Given our knowledge of the original reference distribution of materials for \textit{Density}, we only need to estimate the weights of the \ac{gmm} using the \ac{em} algorithm. This approach allows the classification of unknown objects into material categories without necessitating a class for every material. Consequently, objects that fall between two classes can be represented, for instance, as a mixture, like $0.5 \times property_1, 0.5 \times property_2$. Subsequently, this message vector from \textit{Elasticity} serves as a virtual measurement for the \textit{Material} node. Posterior probabilities are computed using a confusion matrix to model the inherent uncertainty. Specifically, this confusion matrix is unique to each connection between nodes; it is determined beforehand at the initialization step from the reference data for every blue arrow in the network model depicted in \figref{fig:probability_model}. 

When the message originates from a categorical property, i.e., a vector, the estimation step is bypassed, and message passing occurs directly from the product of the measurement vector and the associated confusion matrix.

\subsection{Measurement Models}
The action selection algorithm needs to estimate the consequences of performing different actions, i.e. measurements. A measurement model specifying the accuracy of every measurement is required. 
%To compute the expected \ac{ig}, some description of the measurement uncertainty is needed. 
For categorical variables, confusion matrices are used to compute the expected measurement. For brevity, only the average accuracy for each categorical action is given. Complete confusion matrices are available in the code repository. For continuous variables, a measurement is modeled as a normal distribution. The measurement accuracies and \acp{sd} are provided in the \tabref{tab:actions_summary}. The \ac{sd} for density was experimentally obtained based on the average performance of extracting a correct density belief from a mass measurement and a volume belief on the available experimental data. The mass measurement with our experimental setup has an accuracy of $\pm$ 6 [grams] for up to a 1 [kg] payload.

\begin{table}[]
\resizebox{\columnwidth}{!}{%
\begin{tabular}{|c|c|c|c|c|}
\hline
\textbf{Action} & \textbf{Target node} & \textbf{Accuracy} & \textbf{\ac{sd} $\sigma$} & \textbf{Units}\\ \hline
mat-vision & Material (categorical) & 0.2431 & - & - \\
mat-sound & Material (categorical) & 0.1002 & - & - \\
cat-vision & Category (categorical)& 0.6345 & - & - \\
weighing & Density (continuous)& - & 223 & [kg$\cdot$m$^{-3}$] \\
squeezing & Elasticity (continuous)& - & 10 & [kPa] \\ \hline
\end{tabular}%
}
\caption{Measurement accuracy. Accuracies and \acfp{sd} from experimental data.}
\label{tab:actions_summary}
\customVspace{-1.5em}
\end{table}

\subsection{Expected Information Gain}
%Information gain is a quality measure based on information theory.
In our framework, \ac{ig} quantifies the reduction in entropy of the probability distribution of one or more variables in our \ac{bn} after executing an action. It reflects how much information the action and its associated measurement have provided. When we compute the difference in entropy before and after conducting an action, we refer to it as \textit{experimental} or \textit{true} \ac{ig}. While experimental \ac{ig} offers insight into the performance of an algorithm, by definition, it alone cannot be used to decide which action to take. Therefore, to guide the decision-making, one must create an emulation---an estimation of the property distribution after each action---to choose the action most likely to reduce entropy, thus increasing \ac{ig}. The experimental \ac{ig} is formalized as:
\begin{equation}
    \Delta H(\nu) = H(\nu_k) - H(\nu_{k+1}),
    \label{eq:experimental_info_gain}
\end{equation}
where $\nu$ represents a distribution of a node, either continuous or categorical (e.g., \textit{Elasticity}: \ac{pdf} or \textit{Category}: \ac{pmf}), 
$\Delta H(\nu)$ denotes the experimental \ac{ig}, which is the difference in entropies. Here, $H(\nu{_k})$ is the entropy of a set of nodes at algorithm iteration $k$, representing the prior, and $H(\nu_{k+1})$ is the entropy of the same set of nodes at the next iteration step $k+1$, representing the posterior after the measurement. The expected \ac{ig} is formalized as follows:

\begin{equation}
    \mathbb{E}[\Delta H(\nu)] = H(\nu_k) - \mathbb{E}[H(\Bar{\nu_k})],
    \label{eq:expected_info_gain}
\end{equation}
where $\mathbb{E}[\Delta H(\nu)]$ is the expected \ac{ig}, and $\mathbb{E}[H(\Bar{\nu_k})]$ is the expected entropy of the posterior distribution, which is the entropy of the emulation. Note that the emulation also contains the index $k$, not $k+1$, to emphasize that it reflects only an expectation of the posterior based on the current belief. Theoretically, the emulation for continuous properties can be achieved through sampling. However, this is computationally prohibitively expensive. Since the variance of the action related to a given property is available, we chose a more efficient option---convolving the continuous prior probability density function of a given property $\nu_k$ with a normal distribution having a zero mean and variance equal to the variance of the action measurement in question $\mathcal{N}(0, \sigma_{a}^2)$. This procedure propagates the expected error across the entire continuous probability density function of the property. Formally:

\begin{equation}
    \Bar{\nu_k} = \nu_k \ast \mathcal{N}(0, \sigma_{a}^2).
    \label{eq:continuous_emulation}
\end{equation}

Subsequently, the expected continuous entropy is numerically approximated as:
\begin{equation}
    \mathbb{E}[H(\Bar{\nu_k})] = -\sum_\mathcal{X} h_{\mathcal{X}} \Bar{\nu_k}(x) \log_2\Bar{\nu_k}(x),
    \label{eq:expected_cont_entropy}
\end{equation}
where the entropy is computed for each step on the support $\mathcal{X}$ of the property distribution. $h_{\mathcal{X}}$ denotes the normalization factor, determined by the bin size used for computing the sum-integral approximation. Strictly mathematically speaking, \equationref{eq:expected_cont_entropy} provides an integral approximation.

To represent the variance of the measurements for categorical properties, such as the \textit{Category vision module}, the confusion matrix is employed. From this confusion matrix, a sample entropy vector is computed to create a categorical equivalent of the emulation in \equationref{eq:continuous_emulation}. Each element of the sample entropy vector at index $i$ is computed as
\begin{equation}
    H(\Bar{\nu_k} | m_i) = -\sum_j C_a(i,j) \log_2 C_a(i,j),
    \label{eq:samp_ent_vector}
\end{equation}
where $C_a$ is a row-unit-sum confusion matrix corresponding to action $a$, $m_i$ signifies the $i$-th member of the categorical property, e.g. \textit{ceramic} for the \textit{Material} node or a \textit{mug} for the \textit{Category} node. In the case of a column-unit-sum confusion matrix, the indices $(i,j)$ in \equationref{eq:samp_ent_vector} would interchange. To compute the expected entropy of the emulation, we use the resultant vector from \equationref{eq:samp_ent_vector} as:

\begin{equation}
    \mathbb{E}_m[H(\Bar{\nu_k})] = \sum_{m_i}C_a\nu_k(m_i) H(\Bar{\nu_k} | m_i),
    \label{eq:categorical_emulation}
\end{equation}
where the result is a sum of the sample entropy vector weighted by an expected measurement computed as a matrix multiplication of the prior distribution and confusion matrix as $C_a\nu_k$. The result \equationref{eq:categorical_emulation} is then substituted into \equationref{eq:expected_info_gain} to obtain the expected \ac{ig}. 

\subsection{Action Selection}
The approach is schematically illustrated in \figref{fig:acsel_loop} and the pseudo-algorithm is shown in \algoref{alg:whole_network}.
%Utilizing the reference distribution of the properties and the initial prior probabilities, we compute the expected \ac{ig} for each action and choose the one offering the highest \ac{ig}.
Formally, the optimal exploratory action is chosen as: 
\begin{equation}
    a^* = \text{argmax}_{a \in \mathcal{A}} \mathbb{E}[\Delta H(\nu | a)],
\end{equation}
where $a$ is a specific action from the total set of available actions $\mathcal{A}$, and $a^*$ is the optimal action that maximizes the expected \ac{ig}.

\begin{algorithm}
\small
\SetAlgoLined

\For{k \textbf{in} iterations}{
    $\eta_k := \text{getNetworkEntropy}(N)$\\
    \For{a $\in \mathcal{A}$}
    {   
        $\nu_{k,a} := \nu_k \in N \; | \;a \rightarrow \nu_k $\\
        $\overline{\nu_k} := \text{emulatePosterior}(\nu_{k,a}, a) $\\
        $\mu_k := \text{generateMessages}(\overline{\nu_k}) $\\
        $\overline{N} := \text{propagateMessages}(N, \mu_k) $\\
        $\overline{\eta_k} := \text{getExpectedNetworkEntropy}(\overline{N}) $\\
        $\mathbb{E}_k[\Delta H(N)] := \eta_k - \overline{\eta_k}$\\
    }
    $a_k^* \in \textit{argmax}_{a \in \mathcal{A}_k} \mathbb{E}_k[\Delta H(N)]$\\
}
\caption{Action selection (ACTSEL)}
\label{alg:whole_network}
\end{algorithm}

\section{Objects, Robot Setup, Action Repertoire}
\label{sec:setup}
%Our experimental setup consists of everyday household objects and the exploration setup. 

The setup and robot actions are illustrated in the accompanying video at \url{https://youtu.be/h_ZIYUmzv-8}.

\subsection{Manipulated Objects}
\begin{figure}[tb]
    \centering
    \includegraphics[width=\columnwidth]{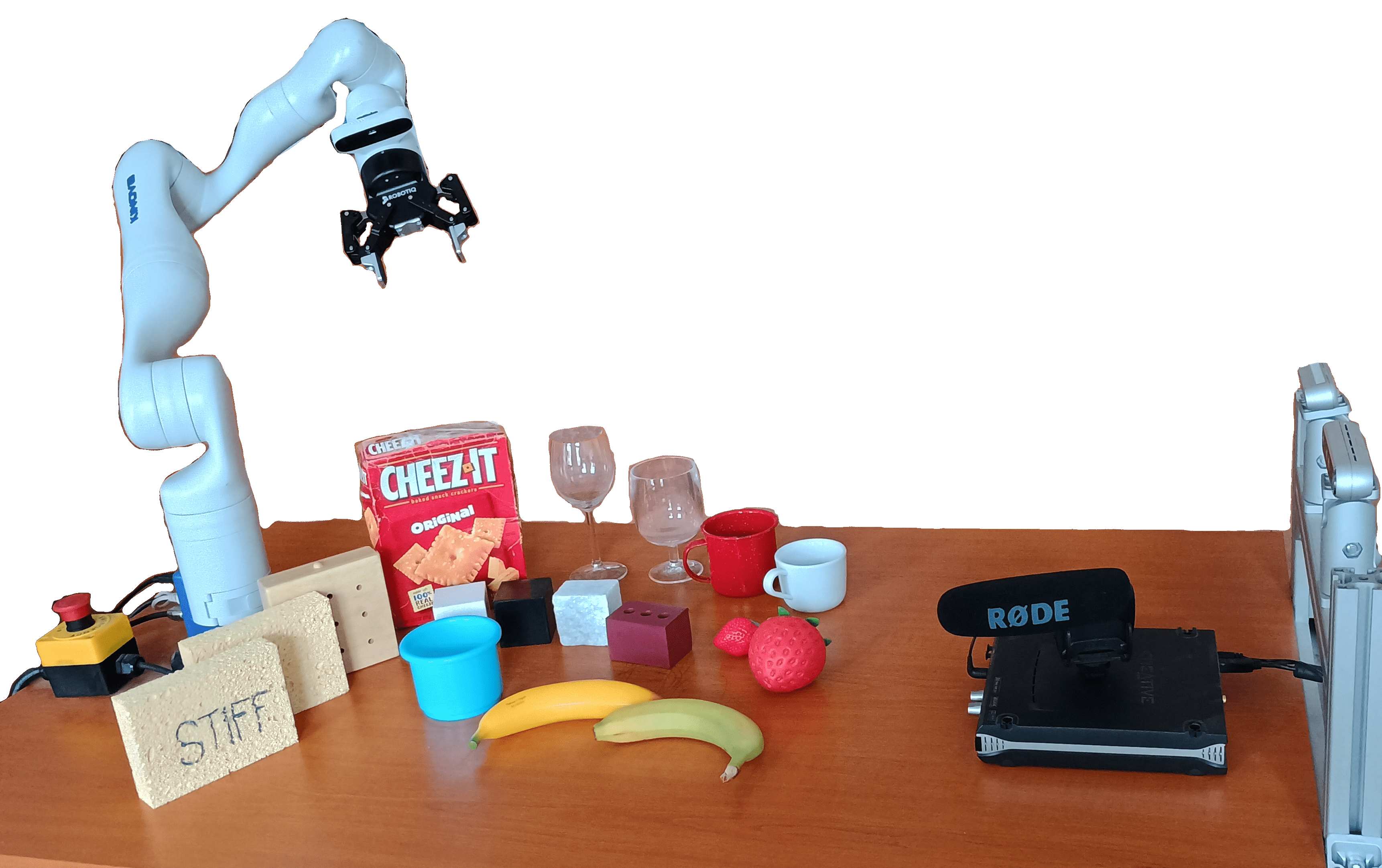}
    \caption{All 17 objects used and experimental setup. Kinova Gen3 robot, Robotiq 2f-85 gripper, Rode VideoMic Pro microphone, and RealSense D435 camera (on the right; only the RGB from one camera was used).}
    \label{fig:setup}
    \customVspace{-1.5em}
\end{figure}

We selected 17 objects, most of them from the YCB dataset~\cite{Calli2015}, to be diverse in size, weight, volume, and  material composition---see \figref{fig:setup}. We also included ``trick'' or adversarial objects where their material composition does not correspond to the best guess from their appearance (real and plastic bananas, a soft sponge and an identical but hardened sponge, wine glass from glass or plastic). 
%The final selection was constrained by the width of the gripper. 

\subsection{Robot Setup}
The real-world robot setup (\figref{fig:setup}) consisted of a Kinova Gen3 robotic manipulator (7 degrees of freedom), equipped with a Robotiq 2f-85 gripper used for \textit{squeezing} and \textit{weighing} actions. In addition, the gripper was used to poke objects to generate sounds for material classification (\textit{mat-sound}). For this purpose, a professional shotgun microphone Rode VideoMic Pro was used. RGB images from a RealSense D435 camera were used for the measurements relying on vision (\textit{cat-vision}, \textit{mat-vision}).

\subsection{Action Repertoire}
\label{subsec:Action_Repertoire}

\textbf{Weigh.} 
The mass of objects is estimated by lifting them using the robot arm using torque measurements in the robot's joints in a configuration where a certain robot joint axis is perpendicular to the force of gravity. The torque in this joint $\tau_m$ can be described by the formula $\tau_m = m  g  l + \tau_0$, where $m$ is the object mass, $l$ is the effective lever from the joint axis to the object center of mass, and $g$ is the gravitation constant, and $\tau_0$ is a reference torque recorded by moving the robot to the measuring pose while the gripper is empty. 

%This method's result depends on an estimate of the effective lever $l$. As this depends on the actual grasp pose, the lever varies with the grasp pose distribution perpendicular to the measuring joint axis. This propagates to the mass estimate. Other influences are covered by the torque reference $\tau_0$. However, $\tau_0$ drifts over time (e.g., due to temperature changes or wear and tear on the robot). How discriminative such a weighing action can finally be depends also on the distribution of the objects and the knowledge about them.

\textbf{Squeeze.}
%This module~\cite{patni2024evaluating} estimates the stiffness of different objects by doing the \textit{squeezing} action and classifies them into different material classes based on the obtained stiffness value, and is focused towards a broad range of soft, deformable objects --- \textit{foams and sponges}; \textit{cardboard and paper} objects like cartons and packing material; \textit{plastic} containers and toys; and recyclable \textit{metal} cans which can be compressed by the robot. Rigid objects that exceed the robots' force handling capacities are reported as \textit{too stiff}.
Grasping or squeezing the object between the jaws of a parallel gripper provides force and displacement data---the stress-strain response of every object. We adapted the module we developed in \cite{patni2024online} and approximated the stress-strain curve with a line---a coarse approximation of the object's stiffness.

%For this paper, we simplify our mechanical model by assuming our stress-strain curve can be fit to a line. While this assumption reduces the accuracy of our estimation, the obtained values are still suitable representations of objects' material properties. An in-depth analysis of different mechanical models used to estimate stiffness via robotic squeezing is provided in~\cite{patni2024evaluating}.
%\href{https://osf.io/gec6s}{https://osf.io/gec6s}

\textbf{Material from sound.}
We used sound generated by poking the object with the closed gripper jaws to recognize the material. Audio recordings from the interactions were preprocessed and converted to grayscale melodic spectrograms and then fed into a modified ResNet34 network, with output labels representing the material classes. More details can be found in \cite{Dimiccoli_2022}.

%We adapted the module developed in  \cite{Dimiccoli_2022} that recognizes the material from which target objects are made by using the audio recordings of robots interacting with these objects as input. For robot-object interaction, we are focusing on the discrete, single-instance action of \textit{poking}. Data collection for training the module was carried out in two different modes --- robot-object interaction with 49 objects from the YCB dataset~\cite{Calli2015}, consisting of regular, repeated poking actions; and human-object interaction with 75 objects from the YCB dataset, consisting of more random approach directions and force ranges. All of this data is available in the \href{https://osf.io/4tcp6}{YCB-impact sounds dataset}. Audio recordings from these interactions were pre-processed and converted to grayscale melodic spectrograms for training a modified ResNet34 network, with output labels representing eleven material classes. We achieved 67\% accuracy for the material recognition task across 11 material classes, and 79\% accuracy for the object recognition task across 75 objects.

\textbf{Material and Category from Vision.} 
%This module is an important component, as it is always used as the first action. \lr{is true? Why is it like that?} \jh{Presumably because it adds the most information - Andrej had priors for each measurement method. Could have been precision? Please clarify Andrej?}
We used the Detectron2 algorithm~\cite{wu2019detectron2} with R50-FPN backbone and trained the model using our dataset consisting of objects from the YCB dataset~\cite{Calli2015}, SHOP-VRB~\cite{nazarczuk2020shop} and our own generated in Unity.\footnote{Dataset available at \href{https://github.com/Hartvi/ipalm-sponges}{https://github.com/Hartvi/ipalm-sponges}. Code available at \href{https://github.com/Hartvi/Detectron2-mobilenet}{https://github.com/Hartvi/Detectron2-mobilenet}.} 
 
%that uses Detectron2 to localize and determine \textit{Material} and \textit{Category}
%The pipeline is available \href{https://github.com/Hartvi/Detectron2-mobilenet}{online}. 

\section{Experiments and Results}
\label{sec:exps}
In our experiments, one of 17 objects was placed in front of the robot. There were 5 actions or measurements (\textit{cat-vision, mat-vision, mat-sound, weighing, squeezing}) in the repertoire that could be selected by our algorithm based on their expected \ac{ig}, i.e. their likelihood to maximally reduce the uncertainty in one or more of the object properties. There were five object properties (2 discrete variables with \ac{pmf} -- \textit{Category} and \textit{Material}; 3 continuous with \ac{pdf} -- \textit{Elasticity}, \textit{Density}, \textit{Volume/Mass})---see \figref{fig:probability_model} for an overview. The \textit{cat-vision} was the action/measurement informing the \textit{Category} node; \textit{mat-vision} or \textit{mat-sound} could be employed to update the \textit{Material} node; \textit{squeezing} measured \textit{Elasticity}. \textit{Density} and \textit{Volume} nodes did not have a corresponding measurement that would update them directly and relied on relationships within the \ac{bn}. 

The actions were selected ``without replacement'', i.e. once a particular action was chosen, it was not available in the next step of the algorithm. Thus, with 5 actions in the repertoire, there were at most 5 algorithm steps. After every measurement, the \ac{bn} was updated as explained in \secref{subsub:inference}. We ran two variants of the experiments: (a) with no stopping criterion---5 actions were always executed; (b) with stopping---if at a given algorithm step, there was no action available with positive expected \ac{ig}, the algorithm would terminate. There were 5 runs of the algorithm for every object. 

\textbf{Optimization modes.} To drive the action selection, one of the nodes could be selected for which the uncertainty should be reduced (say the user is interested in a specific object property). For \textit{Category} or \textit{Material}, the expected discrete \ac{ig} for every action was evaluated to select the best action. For the continuous variables (\textit{Elasticity, Density, Volume}), the corresponding differential \acp{ig} were evaluated. We have also assessed runs optimizing \textit{Category} and \textit{Material} together or the three continuous property nodes together. Aiming to reduce uncertainty in all five nodes at the same time would require calculating a weighted sum of the discrete and differential \ac{ig}, which is not mathematically rigorous, so this regime was not evaluated. 
Please note that all actions have their associated measurement models, i.e. the system has an estimate of the noise associated with every action. Thus, optimizing one property does not always lead to executing the action that directly measures that property. 

\subsection{Qualitative Illustration of the System's Operation}
\begin{figure*}[!htb]
\includegraphics[width=\textwidth]{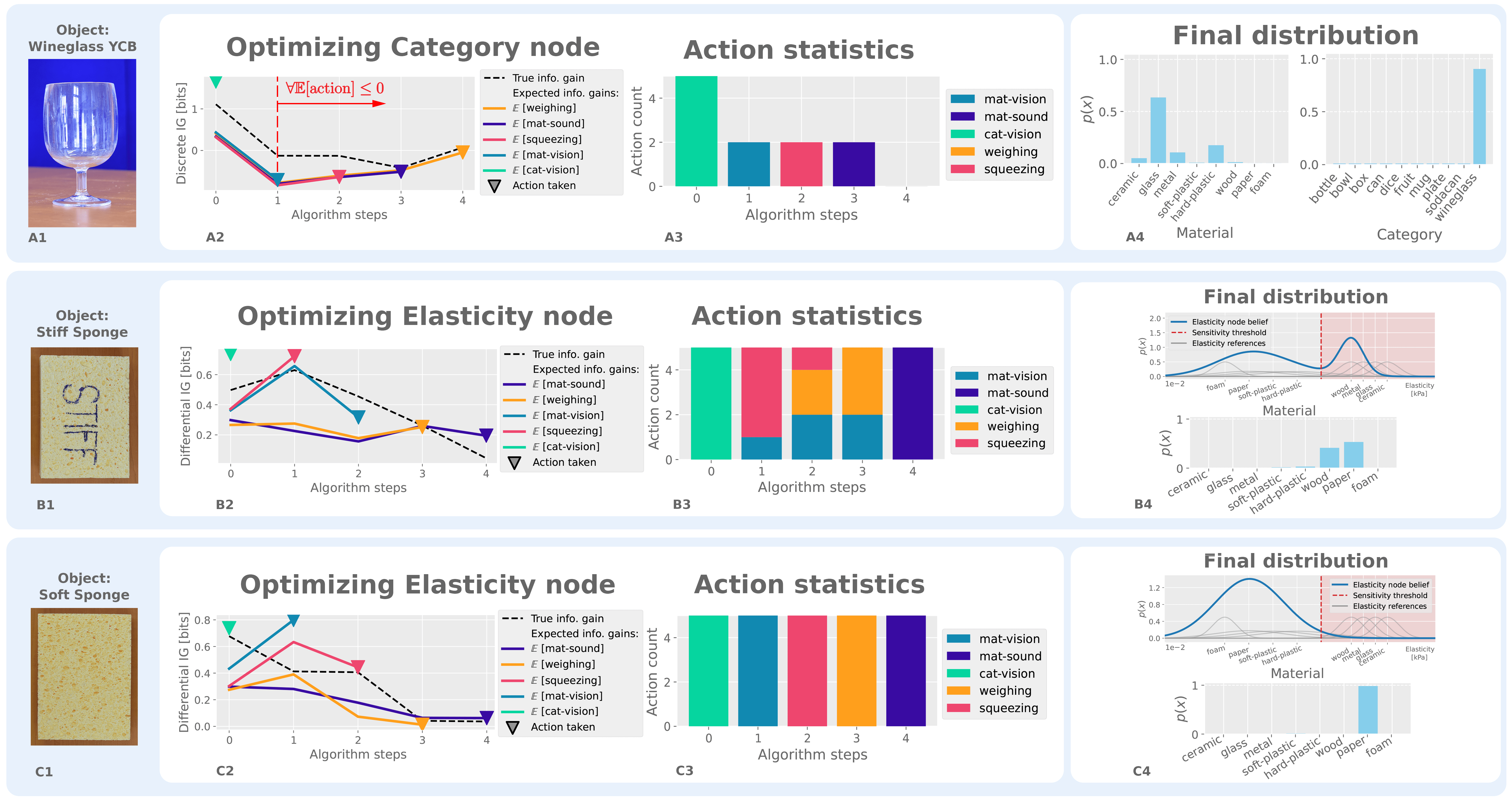}
\caption{System operation and performance -- examples. (Left -- A1/B1/C1) Object name and image. (A2/B2/C2) Single run of experiment with termination criterion off. Expected \ac{ig} of the actions shown. Triangle markers show which action was chosen in a given step. The dashed black line shows the experimental \textit{true} \ac{ig} after the action was executed. This action is not shown anymore in subsequent algorithm steps. (A3/B3/C3) Count of actions chosen during the runs with the termination on (if all available actions at a given step had negative expected \ac{ig}, execution stopped). (A4/B4/C4) Final probability distributions of relevant variables of the single run shown in (A2/B2/C2).}
\label{fig:results_comics}
\customVspace{-1.5em}
\end{figure*}

The operation of the system is illustrated in \figref{fig:results_comics} on three selected objects. In the first row (Panels A1-A4), the \textit{YCB wineglass}, which is plastic, is explored while the \textit{Category} variable is optimized. \textit{Cat-vision} has the highest expected \ac{ig} (panel A2) and is always chosen first (panel A3). In the single run (panel A2), additional steps of the algorithm are shown. However, in this particular run, if termination was on, no more action would be executed as all actions have negative expected \ac{ig} at step 1. Panel A4, right, shows the correct identification of the object \textit{Category} as \textit{wineglass}.

The other two rows show exploration of a sponge---a hardened exemplar in the middle (B1-B4) and a soft one at the bottom (C1-C4) (see accompanying video at \url{https://youtu.be/h_ZIYUmzv-8}); in both cases with learning about \textit{Elasticity} as the objective.  Although the goal was to learn about the object's stiffness, \textit{Cat-vision} was chosen as the first most informative action to serve this objective. The squeezing action executed in the next step (see B2 and B3) adds information about the object being stiff, as is apparent in the final elasticity distribution as well as the material distribution (B4 vs. C4). 

\subsection{Action Statistics and Intelligent Exploration Termination}
\begin{figure}[tb]
\includegraphics[width=\columnwidth]{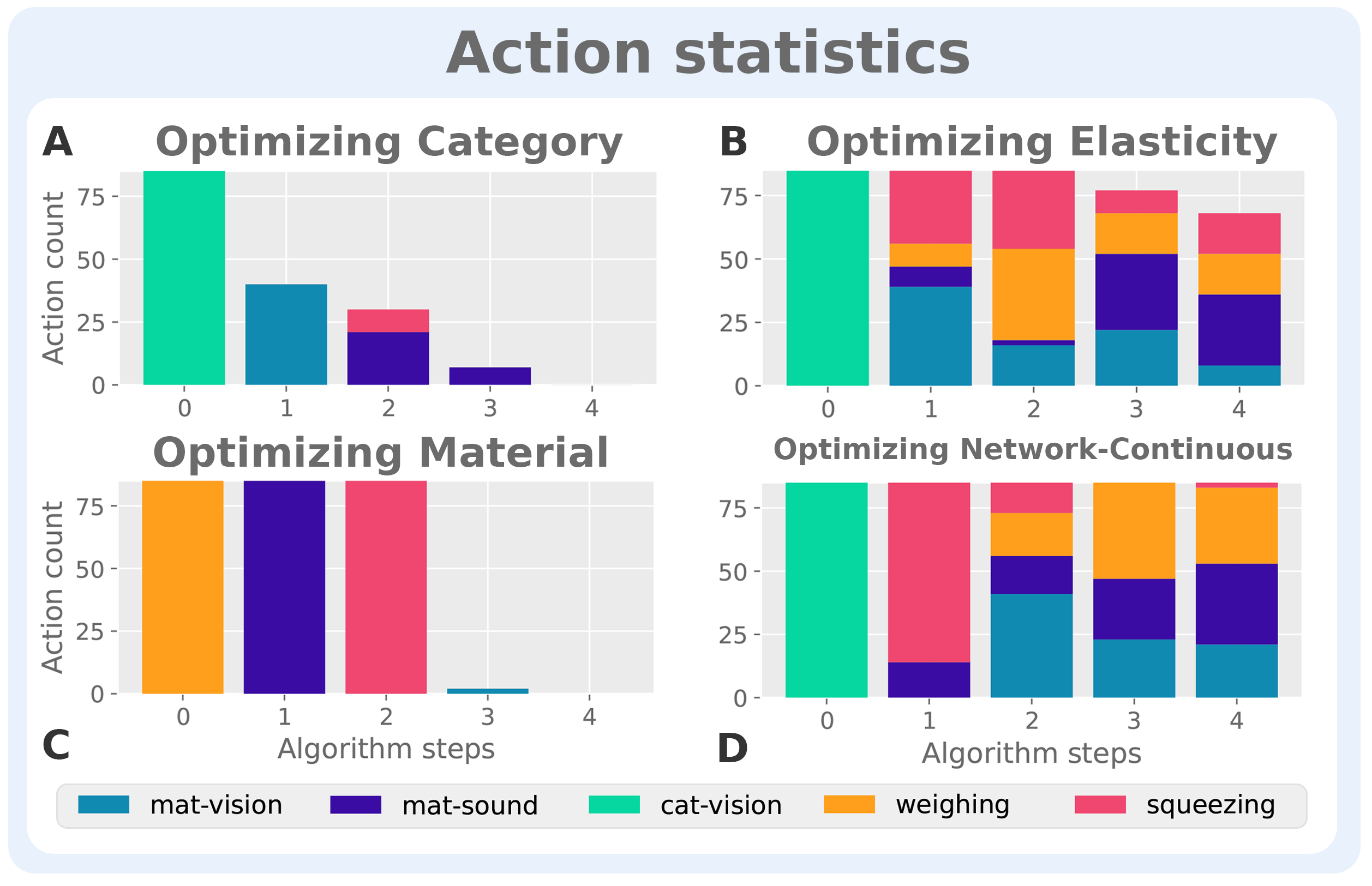}
\caption{Action statistics. Running the algorithm in the active termination mode across all objects, 5 repetitions each. The bar plots show the count of actions selected at every step. Shorter bars indicate that some runs terminated as there was no action with positive expected \ac{ig}. \textit{Optimizing Network-Continuous} jointly optimizes all the continuous variables in the \ac{bn}.}
\label{fig:results_action_stats}
\customVspace{-2.5em}
\end{figure}

\figref{fig:results_action_stats} shows how many times different actions were executed when different object properties, i.e. different nodes in the \ac{bn}, were selected to be ``learned about''. When reducing the uncertainty in the \textit{Category} node, \textit{cat-vision} was always selected as the first action (Panel A). In about half the cases, the algorithm then terminated. In the other cases, \textit{mat-vision} was picked in the second step. When asked to learn about the \textit{Material} (Panel C), interestingly, the algorithm picked the \textit{Weighing} action as the most effective in all cases, rather than the \textit{mat-vision} or \textit{mat-sound} actions that directly target the \textit{Material} node through their measurement update. This illustrates how the algorithm automatically takes the complete \ac{bn} and the measurement uncertainties into account. For the continuous variables (Panels B, D), the algorithm typically performed all five actions. Cat-vision wass the most informative action to learn about \textit{Category} (Panel A), \textit{Elasticity} (B), and all the continuous together variables (D) but not about \textit{Material} (C).

\subsection{Average performance across all objects}
\begin{figure}[tb]
\includegraphics[width=\columnwidth]{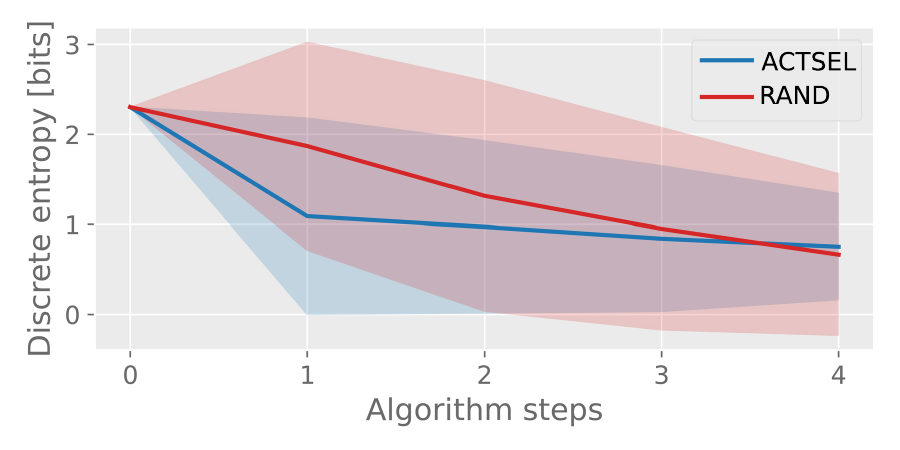}
\caption{Average entropy (lower is better) of the \textit{Category} node -- Optimizing Category, 17 objects, 5 repetitions each. Comparing action selection based on \ac{ig} (\textit{ACTSEL}) and random action selection (\textit{RAND}).}
\label{fig:results_entropy_average}
\customVspace{-1.5em}
\end{figure}

To demonstrate the effectiveness of the proposed algorithm over all the objects, we ran the algorithm with 5 repetitions on each of the 17 objects, using the action selection algorithm proposed optimizing the \textit{Category} node and with random action selection as a baseline (i.e., 2 action selection regimes x 17 objects x 5 repetitions x 5 algorithm steps). For comparison purposes, the termination check was off. \figref{fig:results_entropy_average} shows the average entropy of the \textit{Category} variable after 5 steps of the algorithm in the two action selection regimes, including the variance. The intelligent action selection (\textit{ACTSEL}) outperforms \textit{RAND}. Furthermore, in the \textit{ACTSEL} regime, the plot shows a dramatic drop in entropy after the execution of the first action/measurement, demonstrating that the most effective action was chosen in the first step for the majority of the objects.

\begin{figure}[tb]
\includegraphics[width=\columnwidth]{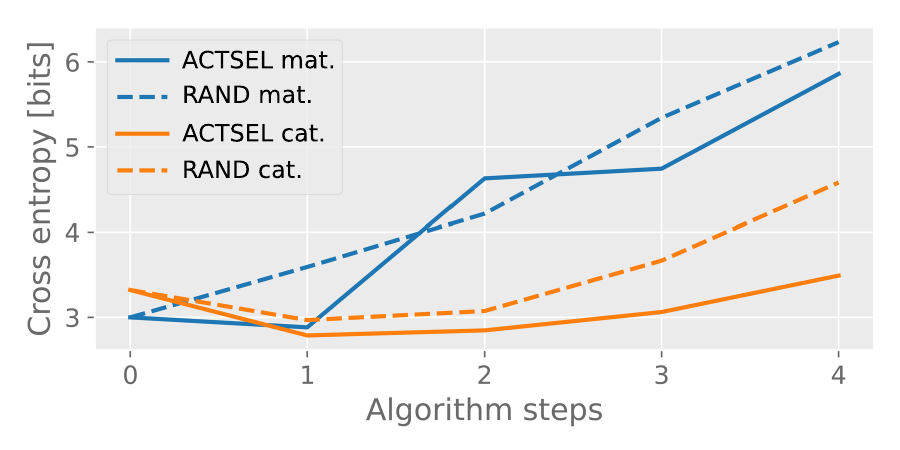}
\caption{Cross-entropy (lower is better) of the \textit{Category} and {Material} nodes, i.e. comparing estimation with ground truth (optimizing Category, 17 objects, 5 repetitions each). Comparing action selection based on \ac{ig} (\textit{ACTSEL}) and random action selection (\textit{RAND}).}
\label{fig:results_cross_entropy}
\customVspace{-1.5em}
\end{figure}

To complement the analysis in \figref{fig:results_entropy_average} that shows entropy, which can be measured autonomously by the system, \figref{fig:results_cross_entropy} displays cross-entropy which considers how well the current probability distribution of a node corresponds to the ground truth---in this case the true label for a given \textit{Category} or \textit{Material}. For both properties, \textit{ACTSEL} on average outperforms \textit{RAND}. The first algorithm step is on average effective in reducing the cross-entropy. Additional measurements may, in fact, add noise to the system, increasing the entropy of the \acp{pdf}. However, in these cases, these steps will often not be executed (see \figref{fig:results_action_stats} A,C).

%In the previous work TODO \cite{andrej's bachelor thesis} \lr{Not sure if citing bachelor theses as previous work is okay}, the algorithm's performance was judged in the sense of a classificator. This was satisfactory for the variable space of the former work, although this work has a significantly higher dimension of the variable space. Some objects' properties do not strictly fall into one or the other category. Therefore, it becomes unclear how to judge the performance of this algorithm. Since the algorithm maximizes the information gain, it minimizes the entropy of the posterior belief. This may seem like a good enough metric on its own, although judging an entropy-based algorithm on the entropy minimization becomes a biased metric. Therefore, we judge the algorithm using cross-entropy. If the algorithm minimizes cross-entropy, that means it minimizes entropy while also minimizing the distance of output labels from the reference labels. As mentioned before, the algorithm's performance is dependent in part on the quality of the actions and whether the provided variance matches the real-world data.

\section{Database of object properties and measurements for robot manipulation}

%\textbf{Datasets of object properties}.
%Among readily available physical object property datasets are the following:
Object datasets related to robot manipulation typically contain object models and visual representations like images, point clouds, or meshes (e.g., YCB \cite{Calli2015}). Datasets containing physical object properties are an exception (e.g., \cite{chen2021comphy,purri2020teaching,thosar2021multi}). 
% \lr{\cite{chen2021comphy} deals with estimating physical properties from contacts from video. The dataset has material, friction, mass}
% \lr{\cite{purri2020teaching} estimation of tactile properties from image "A total of fifteen tactile physical properties across categories including friction, compliance, adhesion, texture, and thermal conductance are measured and then estimated by our models."}

%\begin{enumerate*}[label=(\roman*)]
%    \item YCB -- meshes, textures, physical object procurement \cite{Calli2015};
%    \item ROS household objects - meshes, textures \cite{ros_household2014};
%    \item OCID -- pointcloud categoric segmentation, images \cite{suchiPFV19};
%    \item ARID -- images, pointclouds, object bounding boxes and their categories \cite{arid};
%    \item Robot-centric dataset -- roughness, stiffness, size, including semantic properties like flatness, hollowness, etc. \cite{thosar2021multi}.
%\end{enumerate*}

\begin{figure}[tb]
\includegraphics[width=\columnwidth]{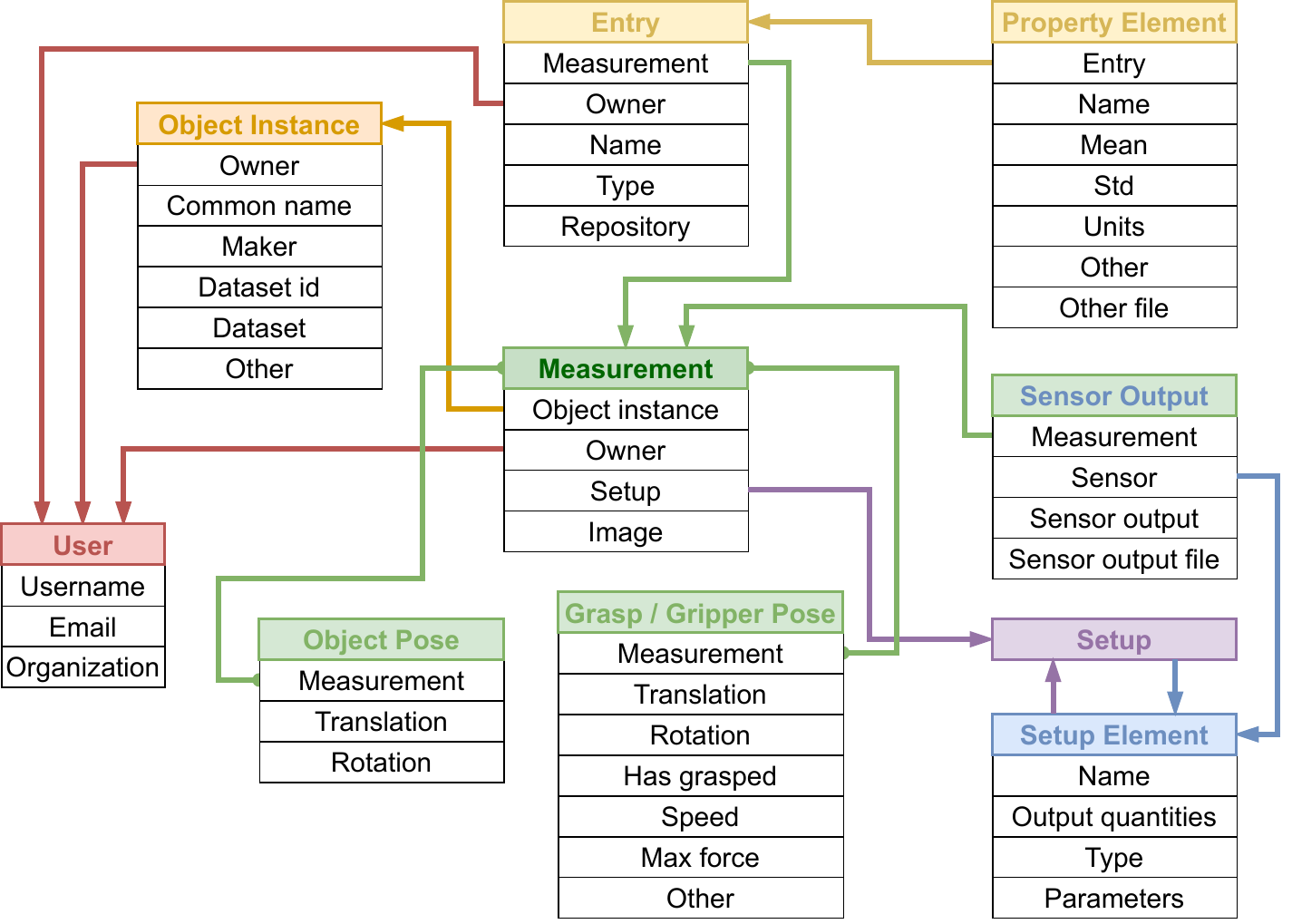}
\caption{Database of object properties and measurements for robot manipulation -- structure.}
\label{fig:the_database}
\customVspace{-1.5em}
\end{figure}

The measurements collected in this work provided the foundation for a database of measurements of object physical properties. These measurements are naturally dependent on the objects manipulated and also highly dependent on the embodiment of the robot (e.g., stiffness measurements on the gripper used and the force/effort feedback)---unlike for images where the camera that recorded the images plays a less critical role. To be useful, such a database needs to cover different setups: different robots, grippers, sensors, etc. and their description in the metadata. The structure of the database we created is shown in \figref{fig:the_database}; it is available at \url{https://cmp.felk.cvut.cz/ipalm/}.

The database contains all the measurements collected in this work (images, stress/strain curves for stiffness estimation, joint angle and joint torque measurements for mass estimation, and audio responses after poking objects). We have already extended the database with elasticity measurements of the same and different objects using several other grippers (OnRobot RG6, Barrett Hand, QB SoftHand, professional elasticity measurement setup). 

Our contribution is aligned with the efforts of others to share robot data across platforms and continents \cite{open_x_embodiment_rt_x_2023}. However, while \cite{open_x_embodiment_rt_x_2023} contains mainly image and joint angle streams, our database is more ``strongly embodied'', allowing for storage of multimodal, in particular haptic data. We invite the community to contribute. To make data collection easier in the correct format, we developed a tool that automates this---\textit{a Python decorator} (\textit{Butler}), and its source code is available on the database website. One can use this decorator to save data in the desired directory structure and upload them to the database. Using this workflow, we populated the database with more than 24000 measurements for over 60 objects. Ultimately, once enough instances of every object type appear in the database, category-level description grounded in the physical measurements will become available. Generalization across the different devices, which were used to collect the haptic data, remains a fundamental challenge.

\section{Conclusion, Discussion, Future Work}
We presented an algorithmic framework and experimental setup for learning physical object properties through robot manipulation. Our approach involved exploratory action selection to maximize expected \acf{ig} about objects on a table in front of a robot manipulator. At its core, the framework employs a \acf{bn} model to represent the probabilistic relationships between different object properties. The \ac{bn} consists of nodes corresponding to various object properties, including categorical properties like \textit{Category} and \textit{Material}, as well as continuous properties such as \textit{Elasticity}, \textit{Density}, and \textit{Volume}.
We demonstrated the effectiveness of our method through experimental evaluation, showcasing its ability to automatically extract properties such as material composition, mass, volume, and stiffness. The algorithm outperformed random action selection baseline and showed promising results in discovering object properties on a partially adversarial object dataset (with ``fake objects'' like a stiff sponge). The algorithm can automatically terminate based on the expected \ac{ig} of the available actions. 
The code used in this work is available at
\url{https://github.com/ctu-vras/actsel}. The object database and source code are available at \url{https://cmp.felk.cvut.cz/ipalm/}.

%\mh{DISCUSSION STARTS HERE}
%The action selection algorithm proposed in this work consists from five nodes, out of which two are categorical (\textit{Category} and \textit{Material}) and three are continuous (\textit{Elasticity}, \textit{Density} and \textit{Volume}). At every passing of the message, the information gathered from an experiment is transferred through the network. There are multiple sources of uncertainty in the network. First is the measurement variance. This variance is natural and accounted for.

Every measurement type in our framework has its own limitations. For example, the Young's modulus for \textit{Elasticity} is limited by the resolution of current flowing through the gripper. There is a threshold above which the gripper cannot provide any specific information on the object deformation. This threshold was experimentally found to be $95$ [kPa]. In such cases, the output is binary, providing a half-open set of moduli from the threshold onward. This is an uninformative measurement over all properties above this experimental threshold. This brings a discontinuous change of elasticity belief distribution.
%In order to make this change as smooth as possible, the arbitrarily large Young's modulus must be chosen appropriately.

Another source of information degradation is the inference of categorical weights from the continuous \acp{gmm} using the \ac{em} algorithm. This stems from uncertainties on the numerical level when the belief distribution mean moved too far from the reference labels. Therefore, the only information that the \ac{em} algorithm possesses at the location of reference labels is from the far tails of the normal distributions. In the process of normalization to unit-sum probabilities from very small numbers, the numerical errors provide a large variance in results. We tried to tackle this problem using Monte Carlo Markov Chains, more specifically Gibbs' sampling method, although this has shown to be too computationally expensive to run in real time.

As part of future work, community engagement and dataset expansion are essential endeavors. By making our dataset and codebase openly accessible, we aim to foster collaboration and knowledge sharing among researchers. We encourage contributions from the community to enrich the dataset with diverse object measurements and annotations, facilitating a broader range of research applications and algorithm evaluations. Additionally, active engagement with the community allows us to gather valuable feedback and insights, guiding the evolution of our framework and ensuring its relevance and usefulness in real-world settings. Together, these efforts contribute to the collective advancement of robot learning and manipulation capabilities, ultimately driving innovation and progress in the field.

Another promising avenue for advancement lies in the development of adaptive exploration strategies. By leveraging techniques from reinforcement learning and advanced decision-making algorithms, robots can autonomously learn and refine their exploration policies over time. Adaptive exploration strategies can dynamically adjust the selection of actions based on the past experience and the current state of uncertainty in object properties. This allows robots to prioritize actions that are more likely to yield informative measurements, leading to more efficient and effective learning of physical object properties. An expert policy will be a more challenging baseline than random action selection.

%%%%%%%%%%%%%%%%%%%%%%%%%%%%%%%%%%%%%%%%%%%%%%%%%%%%%%%%%%%%%%%%%%%%%%%%%%%%%%%%
\bibliographystyle{IEEEtran}
\bibliography{InteractiveLearningObjectProperties}

\end{document}

%% file: acronyms.tex
\newacro{dnn}[DNN]{Deep Neural Network}
\newacro{fcn}[FCN]{Fully Convolutional Network}
\newacro{pc}[PC]{Point Cloud}
\newacro{sdf}[SDF]{Signed Distance Function}
\newacro{cnn}[CNN]{Convolutional Neural Network}
\newacro{gnn}[GNN]{Graph Neural Network}
\newacro{dl}[DL]{Deep Learning}
\newacro{ml}[ML]{Machine Learning}
\newacro{gpis}[GPIS]{Gaussian Process Implicit Surface}
\newacro{mc}[MC]{Monte Carlo}
\newacro{mlp}[MLP]{Multi-Layer Perceptron}
\newacro{bpa}[BPA]{Ball Pivoting Algorithm}
\acrodefplural{gpis}[GPIS's]{Gaussian Process Implicit Surfaces}
\newacro{gpisp}[GPISP]{Gaussian Process Implicit Shape Potential}
\newacro{gp}[GP]{Gaussian Process}
\acrodefplural{gp}[GPs]{Gaussian Processes}
\newacro{ros}[ROS]{Robot Operating System}
\newacro{icp}[ICP]{Iterative Closest Point}
\newacro{fps}[FPS]{Farthest Point Sampling}
\newacro{igr}[IGR]{Implicit Geometric Regularization for Learning Shapes}
\newacro{js}[JS]{Jaccard similarity}
\newacro{cd}[CD]{Chamfer distance}
\newacro{nerf}[NeRF]{Neural Radiance Fields}
\newacro{gmm}[GMM]{Gaussian Mixture Model}
\acrodefplural{gmm}[GMMs]{Gaussian Mixture Models}
\newacro{sd}[SD]{Standard Deviation}
\newacro{em}[EM]{Expectation–maximization}
\newacro{bn}[BN]{Bayesian Network}
\newacro{pdf}[PDF]{Probability Density Function}
\newacro{pmf}[PMF]{Probability Mass Function}
\newacro{ig}[IG]{Information Gain}

%% file: macro.tex
\newcommand{\equationref}[1]{\hyperref[#1]{Eq.~\eqref{#1}}}
\newcommand{\figref}[1]{\hyperref[#1]{Fig.~\ref*{#1}}}
\newcommand{\tabref}[1]{\hyperref[#1]{Table~\ref*{#1}}}
\newcommand{\secref}[1]{\hyperref[#1]{Section~\ref*{#1}}}
\newcommand{\algoref}[1]{\hyperref[#1]{Alg.~\ref*{#1}}}

\newif\ifuseVspace
\newcommand{\customVspace}[1]{\ifuseVspace \vspace{#1} \fi}

%% file: Figures/algorithm_diagram_2023.pdf_tex
%% Creator: Inkscape 1.3.2 (091e20ef0f, 2023-11-25), www.inkscape.org
%% Accompanies image file 'algorithm_diagram_2023.pdf' (pdf, eps, ps)
\begingroup%
  \makeatletter%
  \providecommand\color[2][]{%
    \errmessage{(Inkscape) Color is used for the text in Inkscape, but the package 'color.sty' is not loaded}%
    \renewcommand\color[2][]{}%
  }%
  \providecommand\transparent[1]{%
    \errmessage{(Inkscape) Transparency is used (non-zero) for the text in Inkscape, but the package 'transparent.sty' is not loaded}%
    \renewcommand\transparent[1]{}%
  }%
  \providecommand\rotatebox[2]{#2}%
  \newcommand*\fsize{\dimexpr\f@size pt\relax}%
  \newcommand*\lineheight[1]{\fontsize{\fsize}{#1\fsize}\selectfont}%
  \ifx\svgwidth\undefined%
    \setlength{\unitlength}{233.04000092bp}%
    \ifx\svgscale\undefined%
      \relax%
    \else%
      \setlength{\unitlength}{\unitlength * \real{\svgscale}}%
    \fi%
  \else%
    \setlength{\unitlength}{\svgwidth}%
  \fi%
  \global\let\svgwidth\undefined%
  \global\let\svgscale\undefined%
  \makeatother%
  \begin{picture}(1,0.91864058)%
    \lineheight{1}%
    \setlength\tabcolsep{0pt}%
    \put(0,0){\includegraphics[width=\unitlength,page=1]{algorithm_diagram_2023.pdf}}%
    \put(0.10406156,0.47784275){\color[rgb]{0.24705882,0.30980392,0.45098039}\makebox(0,0)[lt]{\lineheight{1.25}\smash{\begin{tabular}[t]{l}\textbf{ACTION}\end{tabular}}}}%
    \put(0.07347923,0.43664244){\color[rgb]{0.24705882,0.30980392,0.45098039}\makebox(0,0)[lt]{\lineheight{1.25}\smash{\begin{tabular}[t]{l}\textbf{SELECTION}\end{tabular}}}}%
    \put(0.53112794,0.63780707){\color[rgb]{0.24705882,0.30980392,0.45098039}\makebox(0,0)[lt]{\lineheight{1.25}\smash{\begin{tabular}[t]{l}\textbf{WEIGH}\end{tabular}}}}%
    \put(0.51341037,0.36629805){\color[rgb]{0.24705882,0.30980392,0.45098039}\makebox(0,0)[lt]{\lineheight{1.25}\smash{\begin{tabular}[t]{l}\textbf{SQUEEZE}\end{tabular}}}}%
    \put(0.49626481,0.08536662){\color[rgb]{0.24705882,0.30980392,0.45098039}\makebox(0,0)[lt]{\lineheight{1.25}\smash{\begin{tabular}[t]{l}\textbf{MATERIAL}\end{tabular}}}}%
    \put(0.0557305,0.06696807){\color[rgb]{0.24705882,0.30980392,0.45098039}\makebox(0,0)[lt]{\lineheight{1.25}\smash{\begin{tabular}[t]{l}\textbf{PRIOR BELIEF}\end{tabular}}}}%
    \put(0.74031738,0.50813724){\color[rgb]{0.24705882,0.30980392,0.45098039}\makebox(0,0)[lt]{\lineheight{1.25}\smash{\begin{tabular}[t]{l}\textbf{MATERIAL}\end{tabular}}}}%
    \put(0.54245078,0.05208231){\color[rgb]{0.24705882,0.30980392,0.45098039}\makebox(0,0)[lt]{\lineheight{1.25}\smash{\begin{tabular}[t]{l}\textbf{SOUND}\end{tabular}}}}%
    \put(0.73081747,0.20021356){\color[rgb]{0.24705882,0.30980392,0.45098039}\makebox(0,0)[lt]{\lineheight{1.25}\smash{\begin{tabular}[t]{l}\textbf{CATEGORY}\end{tabular}}}}%
    \put(0.77086585,0.16831674){\color[rgb]{0.24705882,0.30980392,0.45098039}\makebox(0,0)[lt]{\lineheight{1.25}\smash{\begin{tabular}[t]{l}\textbf{VISION}\end{tabular}}}}%
    \put(0.77818498,0.47766944){\color[rgb]{0.24705882,0.30980392,0.45098039}\makebox(0,0)[lt]{\lineheight{1.25}\smash{\begin{tabular}[t]{l}\textbf{VISION}\end{tabular}}}}%
    \put(0,0){\includegraphics[width=\unitlength,page=2]{algorithm_diagram_2023.pdf}}%
  \end{picture}%
\endgroup%

%% file: InteractiveLearningObjectProperties.bbl
\begin{thebibliography}{10}
\providecommand{\url}[1]{#1}
\csname url@rmstyle\endcsname
\providecommand{\newblock}{\relax}
\providecommand{\bibinfo}[2]{#2}
\providecommand\BIBentrySTDinterwordspacing{\spaceskip=0pt\relax}
\providecommand\BIBentryALTinterwordstretchfactor{4}
\providecommand\BIBentryALTinterwordspacing{\spaceskip=\fontdimen2\font plus
\BIBentryALTinterwordstretchfactor\fontdimen3\font minus
  \fontdimen4\font\relax}
\providecommand\BIBforeignlanguage[2]{{%
\expandafter\ifx\csname l@#1\endcsname\relax
\typeout{** WARNING: IEEEtran.bst: No hyphenation pattern has been}%
\typeout{** loaded for the language `#1'. Using the pattern for}%
\typeout{** the default language instead.}%
\else
\language=\csname l@#1\endcsname
\fi
#2}}

\bibitem{behrens2021embodied}
J.~K. Behrens, M.~Nazarczuk, K.~Stepanova, M.~Hoffmann, Y.~Demiris, and
  K.~Mikolajczyk, ``Embodied reasoning for discovering object properties via
  manipulation,'' in \emph{2021 IEEE International Conference on Robotics and
  Automation (ICRA)}.\hskip 1em plus 0.5em minus 0.4em\relax IEEE, 2021, pp.
  10\,139--10\,145.

\bibitem{dutta2023push}
A.~Dutta, E.~Burdet, and M.~Kaboli, ``Push to know!-visuo-tactile based active
  object parameter inference with dual differentiable filtering,'' in
  \emph{2023 IEEE/RSJ International Conference on Intelligent Robots and
  Systems (IROS)}.\hskip 1em plus 0.5em minus 0.4em\relax IEEE, 2023, pp.
  3137--3144.

\bibitem{le2021probabilistic}
T.~N. Le, F.~Verdoja, F.~J. Abu-Dakka, and V.~Kyrki, ``Probabilistic surface
  friction estimation based on visual and haptic measurements,'' \emph{IEEE
  Robotics and Automation Letters}, vol.~6, no.~2, 2021.

\bibitem{xu2013tactile}
D.~Xu, G.~E. Loeb, and J.~A. Fishel, ``{Tactile identification of objects using
  Bayesian exploration},'' in \emph{2013 IEEE International Conference on
  Robotics and Automation}.\hskip 1em plus 0.5em minus 0.4em\relax IEEE, 2013,
  pp. 3056--3061.

\bibitem{murali2022empirical}
P.~K. Murali, R.~Dahiya, and M.~Kaboli, ``An empirical evaluation of various
  information gain criteria for active tactile action selection for pose
  estimation,'' in \emph{2022 IEEE International Conference on Flexible and
  Printable Sensors and Systems (FLEPS)}.\hskip 1em plus 0.5em minus
  0.4em\relax IEEE, 2022, pp. 1--4.

\bibitem{li2020review}
Q.~Li, O.~Kroemer, Z.~Su, F.~F. Veiga, M.~Kaboli, and H.~J. Ritter, ``A review
  of tactile information: Perception and action through touch,'' \emph{IEEE
  Transactions on Robotics}, vol.~36, no.~6, pp. 1619--1634, 2020.

\bibitem{petrovskaya2016active}
A.~Petrovskaya and K.~Hsiao, ``Active manipulation for perception,''
  \emph{Springer Handbook of Robotics}, pp. 1037--1062, 2016.

\bibitem{chu2013using}
V.~Chu, I.~McMahon, L.~Riano, C.~G. McDonald, Q.~He, J.~M. Perez-Tejada,
  M.~Arrigo, N.~Fitter, J.~C. Nappo, T.~Darrell, \emph{et~al.}, ``Using robotic
  exploratory procedures to learn the meaning of haptic adjectives,'' in
  \emph{2013 IEEE International Conference on Robotics and Automation}.\hskip
  1em plus 0.5em minus 0.4em\relax IEEE, 2013, pp. 3048--3055.

\bibitem{fishel2012bayesian}
J.~A. Fishel and G.~E. Loeb, ``{Bayesian Exploration for Intelligent
  Identification of Textures},'' \emph{Frontiers in neurorobotics}, vol.~6,
  p.~4, 2012.

\bibitem{nikandrova2014towards}
E.~Nikandrova, J.~Laaksonen, and V.~Kyrki, ``Towards informative sensor-based
  grasp planning,'' \emph{Robotics and Autonomous Systems}, vol.~62, no.~3, pp.
  340--354, 2014.

\bibitem{mahendran_1996}
M.~Mahendran, ``{The Modulus of Elasticity of Steel - Is It 200 GPa?}''
  \emph{International Specialty Conference on Cold-Formed Steel Structures},
  vol.~5, Oct 1996.

\bibitem{dondi1999}
M.~Dondi, E.~G., M.~M., M.~C., and C.~Mingazzini, ``The chemical composition of
  porcelain stoneware tiles and its influence on microstructure and mechanical
  properties,'' \emph{InterCeram: International Ceramic Review}, vol.~48, pp.
  75--83, 01 1999.

\bibitem{HULAN2020}
T.~Húlan and I.~Štubňa, ``Young's modulus of kaolinite-illite mixtures
  during firing,'' \emph{Applied Clay Science}, vol. 190, p. 105584, 2020.

\bibitem{giancoli_2014}
D.~C. Giancoli, \emph{{Physics: Principles with Applications}}.\hskip 1em plus
  0.5em minus 0.4em\relax Pearson, 2014.

\bibitem{Calli2015}
B.~Calli, A.~Singh, A.~Walsman, S.~Srinivasa, P.~Abbeel, and A.~M. Dollar,
  ``{The YCB object and Model set: Towards common benchmarks for manipulation
  research},'' in \emph{2015 International Conference on Advanced Robotics
  (ICAR)}, 2015, pp. 510--517.

\bibitem{patni2024online}
S.~P. Patni, P.~Stoudek, H.~Chlup, and M.~Hoffmann, ``Online elasticity
  estimation and material sorting using standard robot grippers,'' \emph{The
  International Journal of Advanced Manufacturing Technology}, vol. 132,
  no.~11, pp. 6033--6051, 2024.

\bibitem{Dimiccoli_2022}
M.~Dimiccoli, S.~Patni, M.~Hoffmann, and F.~Moreno-Noguer, ``Recognizing object
  surface material from impact sounds for robot manipulation,'' in \emph{2022
  IEEE/RSJ International Conference on Intelligent Robots and Systems
  (IROS)}.\hskip 1em plus 0.5em minus 0.4em\relax IEEE, 2022, pp. 9280--9287.

\bibitem{wu2019detectron2}
Y.~Wu, A.~Kirillov, F.~Massa, W.-Y. Lo, and R.~Girshick, ``Detectron2,''
  \url{https://github.com/facebookresearch/detectron2}, 2019.

\bibitem{nazarczuk2020shop}
M.~Nazarczuk and K.~Mikolajczyk, ``{SHOP-VRB: A Visual Reasoning Benchmark for
  Object Perception},'' \emph{International Conference on Robotics and
  Automation (ICRA)}, 2020.

\bibitem{chen2021comphy}
Z.~Chen, K.~Yi, Y.~Li, M.~Ding, A.~Torralba, J.~B. Tenenbaum, and C.~Gan,
  ``Comphy: Compositional physical reasoning of objects and events from
  videos,'' in \emph{International Conference on Learning Representations},
  2022.

\bibitem{purri2020teaching}
M.~Purri and K.~Dana, ``Teaching cameras to feel: Estimating tactile physical
  properties of surfaces from images,'' in \emph{European Conference on
  Computer Vision}.\hskip 1em plus 0.5em minus 0.4em\relax Springer, 2020, pp.
  1--20.

\bibitem{thosar2021multi}
M.~Thosar, C.~A. Mueller, G.~J{\"a}ger, J.~Schleiss, N.~Pulugu,
  R.~Mallikarjun~Chennaboina, S.~V. Rao~Jeevangekar, A.~Birk, M.~Pfingsthorn,
  and S.~Zug, ``{From Multi-Modal Property Dataset to Robot-Centric Conceptual
  Knowledge About Household Objects},'' \emph{Frontiers in Robotics and AI},
  vol.~8, p.~87, 2021.

\bibitem{open_x_embodiment_rt_x_2023}
{Open X-Embodiment Collaboration}, A.~Padalkar, \emph{et~al.}, ``Open
  {X-E}mbodiment: Robotic learning datasets and {RT-X} models,''
  \url{https://arxiv.org/abs/2310.08864}, 2023.

\end{thebibliography}
